\documentclass{article}
\usepackage{ijcai24}

\usepackage{times}
\usepackage{soul}
\usepackage{url}
\usepackage[colorlinks = True, linkcolor = red, citecolor = blue]{hyperref}       %
\usepackage[utf8]{inputenc}
\usepackage[small]{caption}
\usepackage{graphicx}
\usepackage{amsmath}
\usepackage{amsthm}
\usepackage{mathtools}
\usepackage{amssymb}
\usepackage{booktabs}
\usepackage{algorithm}
\usepackage{algorithmic}
\usepackage[switch]{lineno}

\usepackage{cleveref}

\title{Challenges in interpretability of additive models}

\author{
Xinyu Zhang$^1$
\and
Julien Martinelli$^2$\and
S.T. John$^{1}$
\affiliations
$^1$Department of Computer Science, Aalto University, Espoo, Finland\\
$^2$Inserm Bordeaux Population Health, Vaccine Research Institute, Université de Bordeaux, Inria Bordeaux Sud-ouest, France\\
}

\begin{document}

\maketitle

\begin{abstract}
    We review generalized additive models as a type of `transparent' model that has recently seen renewed interest in the deep learning community as \emph{neural additive models}. We highlight multiple types of nonidentifiability in this model class and discuss challenges in interpretability, arguing for restraint when claiming `interpretability' or `suitability for safety-critical applications' of such models.
\end{abstract}

\newcommand*{\textcite}[1]{\cite{#1}}
\section{Introduction}

Machine-learning models provide predictions just based on existing training data, without any requirement for understanding. However, understanding how a model arrives at its predictions may be required, for example by law, in high-stakes domains such as healthcare, criminal justice, or finance. This has given rise to ``interpretable'' or ``explainable'' machine learning. However, as pointed out by \textcite{lipton2017mythos}, model interpretability ``is not a monolithic concept'' and requires careful consideration of what exactly we mean by it.

Instead of relying on post-hoc explanation methods that only approximate a model's behavior (e.g.~LIME~\cite{ribeiro2016why} or SHAP~\cite{lundberg2017unified}), it has become popular to develop ``transparent'' or ``glass-box'' models that are ``readily understandable'' \cite{rudin2019stop,rai_explainable_2020}.

Here we focus on (generalized) additive models (GAMs) as a subclass of ``transparent'' models. Additive models have a long history in statistical modeling \cite{hastie1986gams}, but received a resurgence in the deep learning community with ``neural additive models'' \cite{agarwal2021neural} that purport to ``combine some of
the expressivity of DNNs with the inherent intelligibility of generalized additive
models''.

In the following, we provide an overview of recent developments in additive models and discuss how, despite their simplistic nature, additive models may not be as interpretable as one might think.

\section{Background on additive models}
\label{sec:additive_models}
\subsection{General definition}

For a $D$-dimensional input $\mathbf{x} = (x_1, x_2, \dots, x_D)$, an additive model takes the following form: 
\begin{equation}
\label{eqn:additive_model}
    f(\mathbf{x}) = \beta_0  + f_1(x_1) + f_2(x_2) + \dots + f_D(x_D) ,
\end{equation}
where the individual components $\{f_d(\cdot)\}_{1\le d\le D}$ are referred to as \emph{shape functions}.
We can model the response variable $y$ in different settings, like classification or regression, as $g(\mathbb{E}[y|\mathbf{x}]) = f(\mathbf{x})$, where $g(\cdot)$ is a link function.

The shape functions can take various forms such as smoothing splines, polynomials, or neural networks. As such they can model highly nonlinear behaviors of each feature while maintaining a relatively simple and `interpretable' structure since each feature is only involved in one term independently of the others. We will return to this in \cref{ssec:how-we-interpret}.

To train the additive model of \cref{eqn:additive_model}, one needs to specify (i) the functional form of the shape functions and (ii) how they are optimized. We cover both points in the following~section. 

\subsection{Literature review: modeling and learning shape functions}

In the first GAMs, the shape functions were constructed with smoothing splines and iteratively trained through a backfitting algorithm~\cite{hastie1986gams}. Within each iteration of backfitting, features are cycled through and shaped from scratch on residuals obtained by fixing other shape functions. More recently, \cite{Lou2012IntelligibleMF} proposed shallow boosted bagged trees as the shape functions and gradient boosting~\cite{Friedman2000GBM} as an alternative fitting method, which gradually expands shape functions instead of learning an entirely new set of shape functions every iteration. \cite{chang2021interpretable} showed that the tree-based models achieved a better balance of feature sparsity, data fidelity and accuracy compared to the spline-based counterparts. 

The Neural Additive Model (NAM) of \cite{agarwal2021neural}, a linear combination of feed-forward networks that are jointly fitted using standard backpropagation, revived additive models in the deep learning community. Since then, numerous extensions of NAMs have been proposed, bringing in ideas like feature selection, feature interactions, and uncertainty estimates. Notably, the recently proposed Bayesian variant of NAMs~\cite{bouchiat2024improving}, LA-NAMs, leverages linearized Laplace-approximated Bayesian inference over subnetworks, enabling principled uncertainty estimation as well as feature selection for both individual features and pairwise interactions. 

\cite{radenovic2022neural} proposed the Neural Basis Model (NBM) as a more parameter-efficient alternative to the NAM. Instead of separate independent neural networks for each input, NBM learns a basis expansion of each input, where the basis functions are shared among shape functions. With a theoretical bound of \(B = O(\log D)\) for the number of bases, NBM effectively reduces the model size while guaranteeing comparable performance to the NAM. 

As an intrinsically probabilistic model,~\cite{duvenaud2011additive} proposed additive Gaussian Processes (GPs) for modeling additive functions, leveraging a novel kernel encoding additive structure of interactions of all orders. The standard method of maximizing marginal likelihood on the training set is used for fitting hyperparameters. \cite{lu2022additive} extended this idea to the orthogonal additive kernel (OAK), thus enabling a more identifiable and low-dimensional interaction selection by imposing orthogonality between shape functions of different orders~\cite{durrande2012anova}. 
    
\subsection{How we interpret additive models}
\label{ssec:how-we-interpret}

One major perk of additive models is the ability to visualize each feature's contribution to the response across the domain, by plotting the corresponding shape function. 
Moreover, classical feature importance metrics can also be used. For instance, 
~\cite{agarwal2021neural} suggested to define the importance score for feature \(x_i\) as \(\frac{1}{N}\sum_{n=1}^N |f_i(x_{n,i}) - \overline{f_i}|\), where $\overline{f_i}=\frac{1}{N} \sum_{n=1}^N f_i(x_{n,i})$, averaged over the entire training set. Alternatively, GAMI-Net~\cite{yang2021gaminet} utilizes the empirical variance of shape functions, \(\frac{1}{N-1} \sum_{n=1}^N f_i^2(x_{n,i})\)~\footnote{In GAMI-Net, all shape functions are centered ($\overline{f_i} =0$) and the overall offset is modeled using an explicit bias term, thus $\frac{1}{N-1} \sum_{n=1}^N\big[f_i(x_{n,i}) - \overline{f_i}\,\big]^2 = \frac{1}{N-1} \sum_{n=1}^Nf_i^2(x_{n,i}) $.}, which can be connected to the variance-based sensitivity analysis techniques widely used for measuring relative importance of (sets of) inputs to a function \(f(\mathbf{x})\). One of the most prevalent importance measures are the Sobol indices, used for assessing feature importance in additive models with orthogonality constraints~\cite{lu2022additive}.

\subsection{Extensions of additive models}
To ensure intelligibility, additive models often integrate inductive biases, such as the absence of feature interactions and smoothness assumption for shape functions. However, strong inductive biases could compromise both accuracy and fidelity, particularly when data exhibits patterns that cannot be solely explained by individual effects, or sharp changes (jumps). Related extensions have been proposed to address these scenarios. 

\subsubsection{Incorporating higher-order interactions}
The performance gap between tree-based additive model and full-complexity models such as random forests suggests a need for incorporating high-order interaction terms~\cite{Lou2012IntelligibleMF} by %
generalizing \cref{eqn:additive_model}  to
\begin{equation}
\label{eqn:additive_model_higher_order}
    f(\mathbf{x}) = \sum_{p \in \mathcal{P}(\{1,\dots,D\})} f_p(\mathbf{x}_p).
\end{equation}
As the number of additive components grows exponentially with interaction orders, this decomposition is commonly restricted to only first- and second-order interactions both for tractability and interpretability, commonly referred to as GA\(^2\)M variants~\cite{lou2013ga2m}. However, even with this restriction, the search space of \(\binom{D}{2}\) feature pairs can still be computationally challenging when dealing with high-dimensional data.

To effectively identify a subset of pairwise interactions, ~\cite{lou2013ga2m} introduced a greedy forward selection strategy called FAST, which first fits the first-order components, then adds the second-order interactions one by one based on a greedy minimization of the residual error. The sparse interaction additive network (SIAN) of \cite{enouen2023sparse} gathers interactions up to a given order using a secant approximation of the Hessian for second-order interactions and a similar approximation of higher-order derivatives for interactions of order more than two. On the other hand, LA-NAM~\cite{bouchiat2024improving} computes mutual information between the last-layer parameters \(\theta_d\) and \(\theta_d'\) of subnetworks \(f_d\) and \(f_d'\) and selects the top-\(k\) highest pairs \((d, d')\). %
Finally, rather than adaptively selecting interactions, a global sparsity constraint can be imposed on the full interaction set using classic methods such as LASSO~\cite{Lin_2006,liu2020ssam}.

\subsubsection{Modeling ``jagged" data} \cite{chang2021interpretable} observed that additive models with overly smooth shape functions, such as the spline-based models or fused LASSO additive models (FLAM)~\cite{petersen2014fused},  fail to capture true patterns when there are sharp jumps in data compared to tree-based variants. The vanilla NAM~\cite{agarwal2021neural} attempted to model more jagged functions by incorporating ``exponential units'' (ExU) activation functions, which can fit even Bernoulli noise. However, follow-up work suggests that, in practice, standard ReLU activation functions appear to perform better \cite{bouchiat2024improving}.
    
\subsubsection{Uncertainty estimation} Most NAM variants lack an inherent framework for epistemic uncertainty, necessitating model-agnostic uncertainty estimation techniques such as ensembles of models. \cite{bouchiat2024improving} extended NAM with Laplace approximation to Bayesian inference to enable a more principled uncertainty estimate. Gaussian process models such as OAK \cite{lu2022additive} naturally provide uncertainty estimates; however, their use of normalizing flow preprocessing makes their uncertainty estimates questionable in practice, as a stationary Gaussian process is fitted in a highly nonlinearly transformed space.
    
\section{Nonidentifiability}

Papers often make fairly strong claims about having identified shape functions and postulating that this is how it is in the real world. However, there are multiple types of nonidentifiability in additive models, possibly causing multiple models that yield the same predictive performance under very different \emph{explanations} of the data. Various solutions have been proposed for obtaining stable parameter estimates with good predictive performance, often claiming to improve interpretability as well. However, it is crucial to question what interpretability means in each use case and whether identifiability truly satisfies this requirement.

\subsection{Nonidentifiability due to only observing sum of terms}

The most salient type of nonidentifiability stems from only observing the sum of additive components rather than each individual component on its own. 
As an illustration, let us consider a two-dimensional additive function:
\begin{equation}
    f(x_1, x_2) = f_1(x_1) + f_2(x_2).
\end{equation}
If we only observe the overall function value \(f(x_1, x_2)\), then for any constant $\Delta$, we get a different, valid decomposition \(g_1(x_1) + g_2(x_2) = (f_1(x_1) + \Delta) + (f_2(x_2) - \Delta)\). 

Such nonidentifiability was shown to produce unstable parameter estimates causing disparate interpretations with different initializations
as well as unnecessarily complicated models when considering interactions, since higher-order terms can absorb the effect from low-order terms. 

Orthogonality constraints were suggested as a solution~\cite{märtens2019decomposing,lu2022additive}. 
\cite{märtens2019decomposing} considered additional covariate information \(\mathbf{x}\) and its interactions with the latent variables \(\mathbf{z}\) in the context of probabilistic dimensionality reduction, by decomposing the mapping into a linear combination of the marginal and interaction effects, while \cite{lu2022additive} focused on encouraging a more parsimonious interaction set for additive GPs. 
In both works, the proposed kernels enforce the integral of each additive component to be zero with respect to the input measure: 
\begin{equation}
\label{eqn:anova_orthogonal}
\int f_p(\mathbf{x}_p)\mathrm{d}\mu({x_i}) = 0 ~~\forall i \in p.
\end{equation}

Similarly, LA-NAM ensures orthogonality in individual effects through the block-diagonal covariance matrix for the approximate posterior over feature subnetworks. 

\subsection{Nonidentifiability due to concurvity/correlated covariates}

Even with orthogonality constraints, nonidentifiability appears when covariates are not independent of each other. This can be either in the form of multicollinearity or of concurvity. While multicollinearity describes a linear relationship between the input features $x_i$, concurvity extends this notion by considering the shape functions $f_i$. This concept effectively captures 
nonlinear dependencies among features, indicating redundancy in the feature set. Formally, following the definition in~\cite{timothy2003concurvity}, 
let \(\mathcal{H} \in \{(f_1, \dots, f_D) | f_d: \mathbb{R} \rightarrow \mathbb{R} \}\) be a family of functions. We say that concurvity occurs between feature $x_i$ and other features if there exists \((h_1, \dots, h_D) \in \mathcal{H}\) such that \(h_i(x_i)\) can be approximated by a combination of other functions, \(h_i(x_i) \approx \sum_{j \neq i} h_j(x_j) \). In this way, the model becomes nonidentifiable, as for any constant $\Delta$, it holds that:
\begin{equation}
    \label{eqn:concurvity}
    \sum_{d=1} ^D f_d(x_d) = f_i(x_i) + \Delta h_i(x_i) + \sum_{j \neq i} [f_j(x_j) - \Delta h_j(x_j)] 
\end{equation}
Concerns regarding concurvity were already raised in the 1990s when smoothing spline-based additive models were introduced~\cite{buja1989smoother}. Using a more flexible function class such as NAMs only worsens the issue. %

In order to address this type of nonidentifiability, several feature selection algorithms controlling for concurvity have been proposed: both the mRMR~\cite{dejay2013mrmre} and HSIC-Lasso~\cite{climente2019blockhisclasso} methods adopt the Hilbert-Schmidt Independence Criterion (HSIC)~\cite{gretton2005hsic} to measure the dependence between two variables. This enables the formulation of scoring criteria in mRMR and the optimization objective in HSIC-Lasso.
Moreover, the recent work by~\cite{siems2023curve} introduced a concurvity regularizer for differentiable additive models, enforcing decorrelation between transformed features by penalizing the average Pearson correlation coefficient computed on training batches. By carefully choosing the regularization strength from the measured concurvity--accuracy trade-off curve, the authors claim that their method can effectively stabilize model parameter estimates without significantly compromising performance.

\subsubsection{Single identifiable model does not guarantee full insight into data under concurvity}
These algorithms return a non-redundant, more parsimonious feature subset with comparable predictive performance to the full model. However, when strong concurvity exists, multiple subsets with equal size may support similar predictive performance, making it difficult to define the ``right'' answer. Let us consider the extreme case with two-dimensional inputs discussed in~\cite{siems2023curve}, where \(x_1\) and \(x_2\) are identical, i.e.~perfectly correlated: the regularized model will predict solely based on either one of them. However, this selection is arbitrary as there is no natural preference: either \(\{x_1\}\) or \(\{x_2\}\) could be a ``right" subset for the model.

\begin{figure}
\includegraphics[width=1\linewidth]{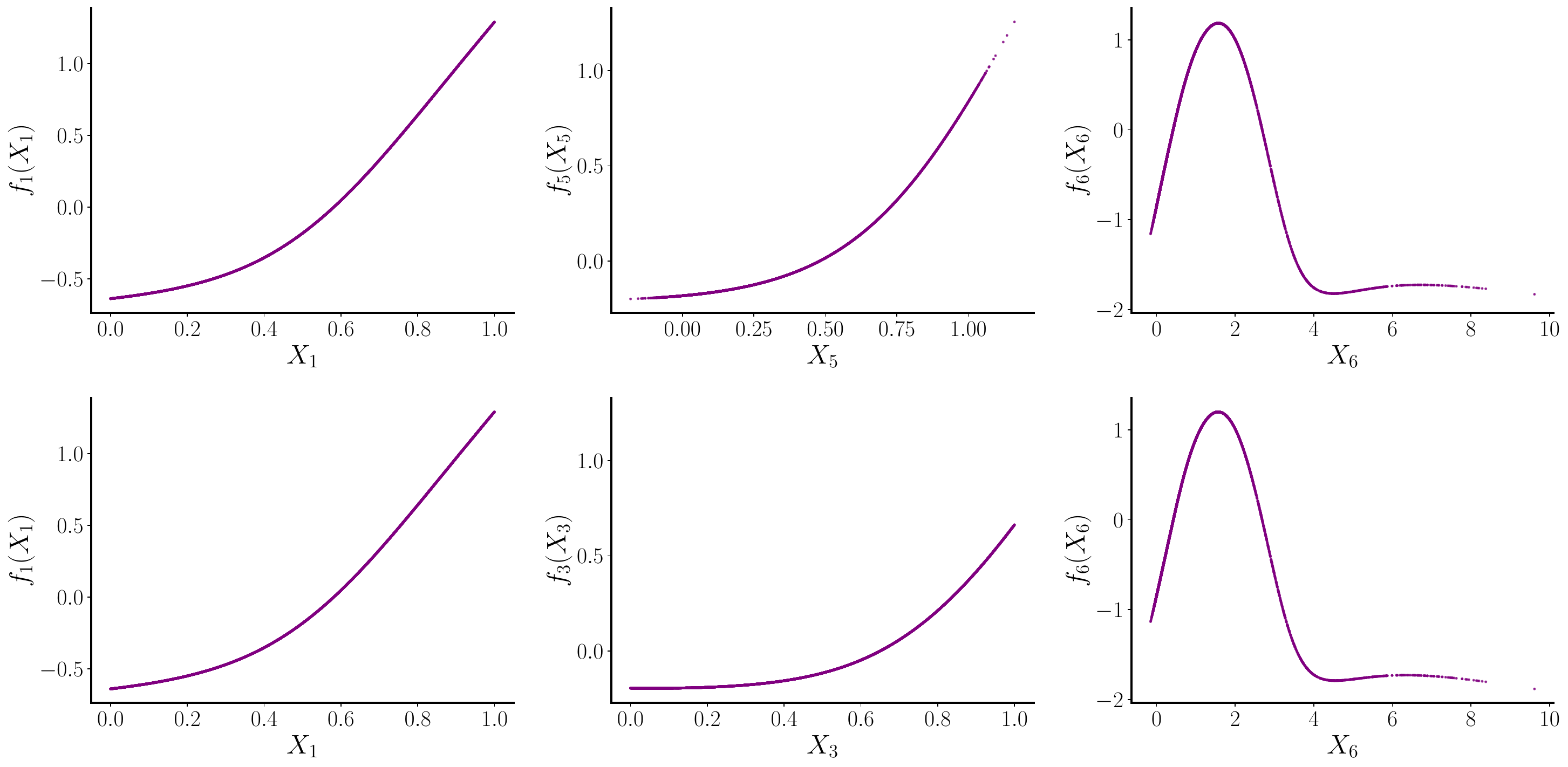}
\caption{Shape functions learned by fitting NAMs on the subsets \(\{X_1,X_5, X_6\}\) (top) and \(\{X_1, X_3, X_6\}\) (bottom) to data generated from \cref{eq:kovacs-example}.}
\label{fig:kovac_shape_functions}
\end{figure}

A more involved example can be found in~\cite{kovacs}: 
\begin{align}
\label{eq:kovacs-example}
\begin{aligned}
    X_1 &\sim X_2 \sim X_3 \sim U(0, 1),\\X_4 &= X_2 ^3 + X_3 ^2 + N(0, \sigma_1),\\X_5 &= X_3 ^2 + N(0, \sigma_1),\\ X_6 &= X_2 ^2 + X_4^2 + N(0, \sigma_1), \\X_7 &= X_1 \cdot X_2 + N(0, \sigma_1),\\Y  &= 2 X_1 ^2 + X_5 ^3 + 2 \sin (X_6 ) + N(0, \sigma_2),
\end{aligned}
\end{align}
with \(\sigma_1 = 0.05, \sigma_2 = 0.5\).
Obtaining nonzero shape functions for \(\{X_1, X_5, X_6\}\) would be regarded as the optimal response as only these features show ``direct'' effects on the target \(Y\).  However, severe concurvity occurs, for example, on feature \(X_5\), as its effect on the target can be easily approximated with the polynomial mapping \(h_3 :X_3 \mapsto X_3 ^6\) (\cref{fig:kovac_shape_functions}). This makes \(\{X_1, X_3, X_6\}\) another reasonable solution, as can be observed from the similar RMSE shown in \cref{tab:kovacs_results}.

This scenario has been described as the ``Rashomon" effect~\cite{breiman2001twocultures}, highlighting the possibility of multiple models giving near-optimal accuracy but different interpretations of data. 
It necessitates a careful specification of the dimension of model interpretability most relevant to each task as emphasized by ~\cite{lipton2017mythos}, such that an appropriate criterion can be determined for meaningful model evaluation. For example, an identifiable algorithm might be sufficient when the main task is to automate the decision-making processes, as predictable model behavior is the primary demand; instead, if the goal is to gain comprehensive insight into the data, relying on one single feature subset supporting the minimal predictive error might not be a safe choice.
\cite{fisher2019models} suggested to analyze variable importance on the \textit{entire \textit{Rashomon set}}, rather than on a \textit{single best} model. Another important line to interpret feature effects is causal inference, which is beyond the scope of our paper.

\begin{table}
    \centering
    \begin{tabular}{cc}
        \toprule
         Input subset & test-set RMSE\\
         \midrule
         Full model & \(0.5205 \pm 0.0096\) \\\ 
         \(\{X_1, X_5, X_6\}\)& \(0.5281 \pm 0.0123\)\\
         \(\{X_1, X_3, X_6\}\)& \(0.5387 \pm0.0092\)\\
         \bottomrule
    \end{tabular}
    \caption{Results of fitting NAMs with different feature sets to data generated from \cref{eq:kovacs-example}. Mean and standard deviation are computed over \(5\) random seeds.}
    \label{tab:kovacs_results}
\end{table}

\subsubsection{Uncorrelated shape functions do not guarantee understandable marginal effects}

\begin{figure}
\includegraphics[width=\linewidth]{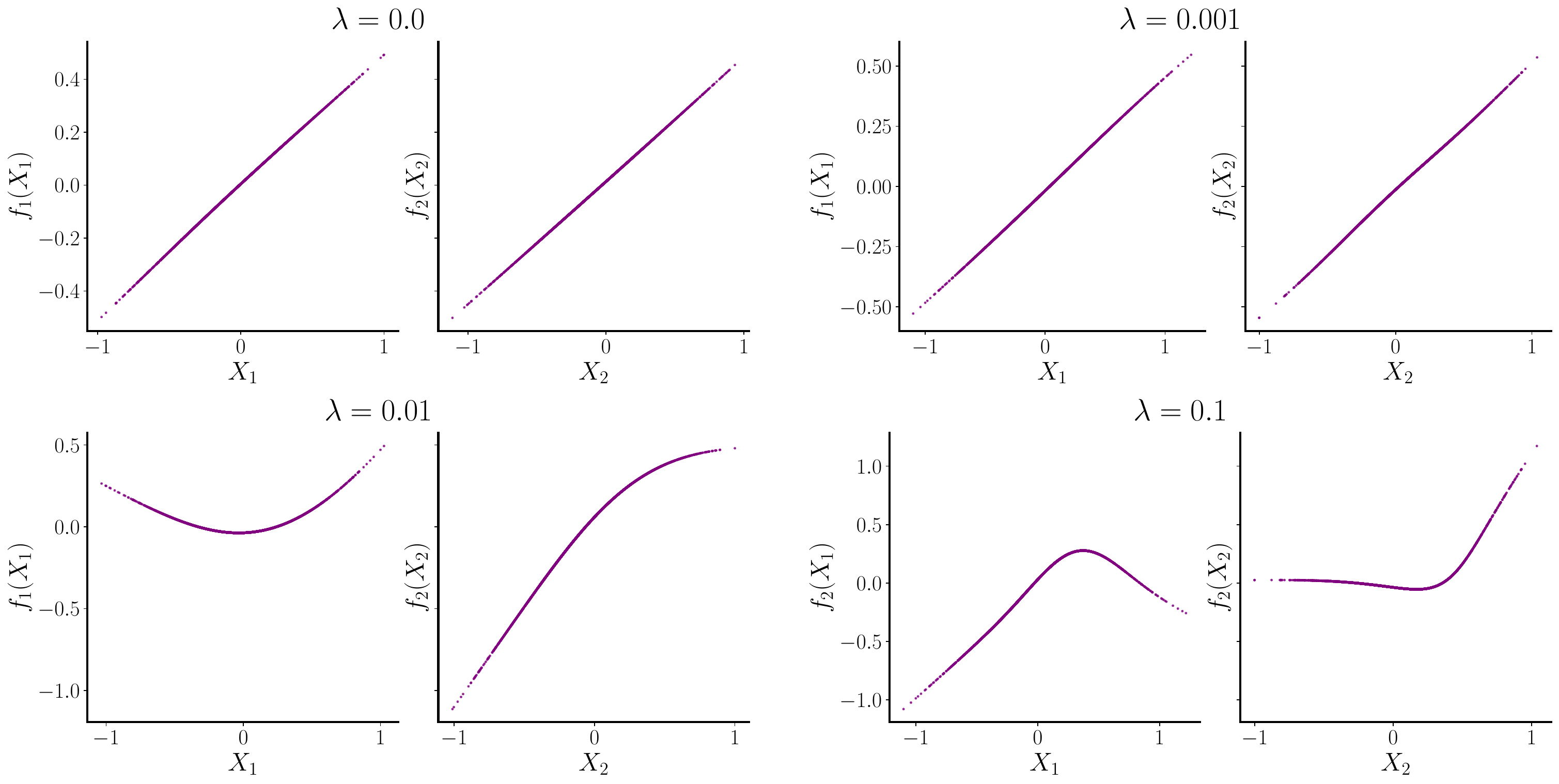}

\caption{Shape functions learned by fitting NAMs with different concurvity regularization strength to data generated from \cref{eqn:interaction-example}.}
\label{fig:interaction_example_shape_functions}
\end{figure}

Let us consider the two-dimensional example
\begin{equation}
\label{eqn:interaction-example}
    f(\mathbf{x}) = \min(x_1, x_2)-0.1x_1 -0.1x_2 ,
\end{equation}
where the individual effects \(f_1(x_1) = -0.1x_1\) and \(f_2(x_2) = -0.1 x_2\) both slope downward, but the interaction effect \(f_{12}(x_1, x_2)\) increases. Variables \(x_1\) and \(x_2\) follow a Gaussian distribution, \((x_1, x_2)^\top \sim \mathcal{N}(\mathbf{0},\Sigma)\). The covariance matrix \(\Sigma\) is represented as \(\begin{bsmallmatrix}1 & \rho \\ \rho & 1\end{bsmallmatrix}\), with an adjustable covariance term. 
We set \(\rho = 0.9\) to induce a strong correlation between \(x_1\) and \(x_2\). Next, we consider the concurvity regularizer introduced by~\cite{siems2023curve} as an additional penalizing term to the optimization objective of NAMs: 
\begin{equation}
\label{eqn:concurvity_opt}
\underset{f_i}{\text{argmin}}~ L(\hat{y}, f_i(x_i)) + \lambda \cdot R_\perp\left(\{f_i\}, \{x_i\}\right).
\end{equation} 
The first term is the task-specific loss function, and \(R_\perp\left(\{f_i\}, \{x_i\}\right)  \) is computed by averaging the Pearson correlation coefficient between shape functions:
\begin{equation}
\label{eqn:concurvity_regularizaion}
    R_\perp \hspace{-.06cm} \left(\{f_i\}, \{x_i\}\right) \hspace{-.06cm} = \hspace{-.06cm}\frac{2}{D(D+1)}\sum_{i=1}^D\sum_{j=i+1}^D \hspace{-.07cm}|\text{Corr}(f_i(x_i), f_j(x_j)| .
\end{equation}
The hyperparameter $\lambda$ governs the tradeoff between goodness-of-fit and the uncorrelated nature of transformed features.
While \cite{siems2023curve} claim that achieving the latter can effectively be seen as a proxy for interpretability,  \cref{fig:interaction_example_shape_functions} shows a different story. Indeed, NAMs fitted using the objective \cref{eqn:concurvity_regularizaion} give rise to gradually less correlated shape functions as $\lambda$ increases. However, they remain dependent in a nonlinear manner. This hinders interpretability, as deciphering the unique contribution of either $x_1$ or $x_2$ on the response remains challenging, and suggests employing other penalties like HSIC to handle nonlinear dependencies.  

\section{Conclusions}

The recent introduction of GAM's deep learning counterpart, NAMs, offered the promise of flexible, easy-to-train and transparent models. While the additive nature of these models can be seen as a ``seal of interpretability'' through their apparent transparency, specifically given the effect of each feature can be visualized, interpretability still remains a challenge. As we illustrated, this is due to several nonidentifiability issues of the shape functions. Though orthogonality constraints effectively allow one to distinguish between $f_1(\cdot) + f_2(\cdot)$ and ($f_1(\cdot)-\Delta)+(f_2(\cdot)-\Delta)$, they fall short as far as concurvity is concerned. As such, linear dependencies can still be found among shape functions. This blurs the interpretation of the actual effect of a feature on the response. Ultimately, since GAMs cannot select optimal feature sets, the selection remains to be done by domain experts. Domain experts may also be able to attribute a \emph{causal} meaning to each feature, thus easing the choice of an optimal feature subset. For this reason, providing experts with an ensemble of solutions rather than a unique candidate set, as proposed by Rashomon-set-based approaches, currently seems like the best tradeoff~\cite{zhong2023exploring}.

\section*{Acknowledgements}
This work was supported by the Research Council of Finland (Flagship programme: Finnish Center for Artificial Intelligence FCAI and decision 341763) and EU Horizon 2020 (European Network of AI Excellence Centres ELISE, grant agreement 951847). We also acknowledge the computational resources provided by the Aalto Science-IT Project from Computer Science IT.
XZ acknowledges the use of OpenAI's GPT-3.5 language model for assistance in polishing the language of this manuscript. The AI tool was utilized to enhance the clarity and readability of the text without influencing the substantive content of the research.

\bibliographystyle{named}
\bibliography{ijcai24}

\end{document}